\documentclass[journal]{IEEEtran}
\usepackage{hyperref} 
\usepackage{times}
\usepackage{epsfig}
\usepackage{graphicx}
\usepackage{amsmath}
\usepackage{amssymb}
\graphicspath{ {images/} }
\usepackage{colortbl}
\usepackage[table]{xcolor}

\ifCLASSINFOpdf
\else
\fi
\begin{document}
%
\title{Long Activity Video Understanding using Functional Object-Oriented Network}
%
%
%

\author{Ahmad~Babaeian~Jelodar,~David~Paulius,~and~Yu~Sun
\thanks{A. B. Jelodar, D. Paulius, and Y. Sun are with the Department
of Computer Science and Engineering, University of South Florida, Florida
FL, 33620 e-mail: (ajelodar@mail.usf.edu).}
}

%
%

\markboth{}%
{Jelodar \MakeLowercase{\textit{et al.}}: Long Activity Video Understanding using Functional Object-Oriented Network}
%



\maketitle

\begin{abstract}
Video understanding is one of the most challenging topics in computer vision.
In this paper, a four-stage video understanding pipeline is presented to simultaneously recognize all atomic actions and the single on-going activity in a video.
This pipeline uses objects and motions from the video and a graph-based knowledge representation network as prior reference.
Two deep networks are trained to identify objects and motions in each video sequence associated with an action. 
Low Level image features are then used to identify objects of interest in that video sequence.
Confidence scores are assigned to objects of interest based on their involvement in the action and to motion classes based on results from a deep neural network that classifies the on-going action in video into motion classes.
Confidence scores are computed for each candidate functional unit associated with an action using a knowledge representation network, object confidences, and motion confidences.
Each action is therefore associated with a functional unit and the sequence of actions is further evaluated to identify the single on-going activity in the video. The knowledge representation used in the pipeline is called the functional object-oriented network which is a graph-based network useful for encoding knowledge about manipulation tasks. 
Experiments are performed on a dataset of cooking videos to test the proposed algorithm with action inference and activity classification.
Experiments show that using functional object oriented network improves video understanding significantly.
\end{abstract}


%
\IEEEpeerreviewmaketitle

\section{Introduction}
\IEEEPARstart{V}{ideo} understanding is a very challenging topic since it would require one to complete several difficult steps successfully, where each step is a challenging and active research topic by itself. 
It would usually require the video to be automatically split into atomic events, the activities and objects in the atomic video clip to be successfully recognized, and a meaningful understanding inferred based on the activities and objects. 
For each step, extensive learning would be carried out for object recognition, activity recognition and video splitting, but they are usually done individually. 

We propose to learn the relationship between objects, activities, and events and represent those relationships in a graph. 
We use the graph as a structured prior information for video understanding when we could. 
For example, a video that demonstrates a chef who is cooking an omelet, comprises of multiple consecutive actions, and each action such as mixing eggs in a bowl deploys multiple objects such as bowl, whisk and eggs. 
To identify the actions, the structural information between the objects (bowl, whisk, egg) and motions (mixing) can be useful. 
For instance, if we understand that eggs can be mixed using a whisk, we could associate the object whisk with the objects egg and bowl. 

The structural information between consecutive actions can be applied to interpret an activity in a video.
For example cracking eggs into a bowl happens before mixing the eggs in the bowl. 
 Consequently, this knowledge can help with predicting that the current on-going action is mixing, knowing that the previous action was cracking eggs.
Embedding these informative structures into a prior graphical structure and using the embedding for inference at test time can improve video understanding.

We use the coordination encoded in the object nodes (e.g. bowl or eggs) and motion nodes (e.g. stirring) of the knowledge-based graph presented in \cite{FOON} to recognize actions (such as stirring eggs) in videos.
The knowledge-based network used for task inference called the {\it functional object-oriented network}, or FOON for short \cite{FOON}, encodes knowledge about the flow of actions coming one after another.
Using this network, we present a powerful object-oriented inference algorithm for action and activity recognition. 

We propose a pipeline that deploys object localities and their motion features to identify active objects within an action. 
We train a deep model for holistic motion recognition which helps with cases where the object (e.g. salt in a chef's hand) is not easily detectable. 
The identified objects and motion are fed to the inference stage with FOON to provide a list of candidate functional units that can be associated with the current on-going action (e.g. cracking egg in a bowl).
The consecutive predicted functional units are evaluated to understand the activity performed in the video (e.g. making Omelet).
This work has four main contributions:
\begin{itemize}
  \item Integrating object localities, object flow features and their accordances in the functional object oriented knowledge representation for action recognition.
  \item Deploying the prior structural information between objects, and motions in the functional object oriented network for functional event recognition in video (e.g. Using the relation between the objects egg, fork and bowl to interpret and label the action as "stirring eggs in a bowl with a fork").
   \item Using the structural information of consecutive actions in the functional object oriented network for task inference (e.g. recipe classification based on a list of consecutively predicted actions).
  \item Merging a deep neural network for motion recognition with the FOON knowledge representation for functional action recognition.
\end{itemize}

The rest of the paper is organized as follows: in Section \ref{pre_work}, we discuss the related work and in Section \ref{section_overall_FOON}, we describe the functional object-oriented network \cite{FOON}.
In Section \ref{section_pipeline}, we introduce the algorithm pipeline. 
In Section \ref{section_FUA}, we explain how objects of interest are identified and
in Section \ref{section_FUR}, we describe the procedure of functional event recognition. 
In Section \ref{section_Exps}, we discuss experiments and results, and we conclude our findings in Section \ref{section_Conclusion}.

\section{Related Work} \label{pre_work}
\subsection{Knowledge Representation} \label{Related_KR}

Knowledge based methods have been successfully applied in natural language processing for Wordnet \cite{WordNet}, Verbnet \cite{VerbNet}, and Framenet \cite{FrameNet}. In \cite{NeverEnding}, Carlson et al. propose a knowledge based architecture to learn a language from web text. 
Some work introduce and use a knowledge base for answering queries \cite{MarkovLogicNetworks}, visual queries \cite{VisualQA2}, and cuisine and ingredient oriented queries using deep features \cite{Review_Reviewer1}.
Knowledge based methods have been also used in visual applications such as the ontological hierarchical knowledge base for image content retrieval and video event detection \cite{Town2006}, scene understanding \cite{ImageInterpret1}, description logics for scene interpretation \cite{LogicDescription}, visual structured knowledge base for scene recognition and object detection \cite{NEIL}, and a combination of various knowledge based representations using machine learning and statistical approaches \cite{Elementary}. 
In \cite{ObjectAffordances_KBR}, the problem of object affordance reasoning is modeled using a knowledge base representation. 
In \cite{Genome} a visual knowledge base representation and dataset is introduced for modeling relationships in images.
In \cite{IEEE_TMM7_KB}, a knowledge representation based method for food recognition from image was proposed which is close to our application.
The lack of a structured knowledge representation as a joint representation of objects and motions motivated us to apply the functional object oriented network for video understanding in cooking videos.

\subsection{Video Understanding} \label{Related_VU}

There is a broad area of work in video understanding. Some works deploy costly setups like physical sensors or additional modalities (e.g. text) \cite{Smart_Kitchen, MultipleSourceSummary, ManipulationPrediction}, some researches perform analysis on spatio-temporal features of a sequence in a holistic manner to label actions \cite{Action_Retreive, CNN+LSTM, Semantic_Model, ImprovedT_Actions}, or use spatio-temporal features of a person (e.g. models of joints or pose) to classify actions \cite{PCNN_Pose,PoseHumanAnalysis}.
These methods are incapable of handling variations in view, zoom and occlusion easily. 
Simultaneous video segmentation and understanding \cite{HierarchichalVideo}, \cite{HMM_CFG}, \cite{Object_access} is also a very common research area.
These methods usually do not consider objects or variations in pose.
Some approaches, extract and analyze a selection of frames for video event summarization \cite{Review_Reviewer2}, and fast anomaly concentration and detection \cite{Review_Reviewer3}.
Jain et al. propose a method that embeds structure into a deep model \cite{CVPR_Review2} to incorporate knowledge with deep models for activity recognition. Other deep approaches proposed for activity recognition are \cite{IEEE_TMM3_Video_deep, IEEE_TMM8_Video_deep}.
A group of research has been conducted to use objects, their affordances, and states in video for action recognition \cite{CVPR_Review3, CVPR_Review4, Scalable, Deep_based, IPM_Mid, ContextModeling}. The motivation to incorporate FOON for video understanding is based on this aspect of activities in video.
Nowadays, there are various multi-view applications especially in surveillance systems.
Information from multiple cameras can enhance event summarization or task understanding.
Various researchers have proposed methods for handling multi-camera scenarios.
Event summarization in multi-view videos using a deep learning approach \cite{multi_view1}, detection and summarization of an event in multi-view surveillance videos by applying boosting \cite{multi_view2}, and a machine learning ensemble method \cite{multi_view3} are instances of research in the area of multi-view video understanding.
We have not addressed this aspect of video understanding in this work.
However, the proposed framework can be deployed in multi-view systems. 
We have a discussion on the multi-view aspects of our methodology in Section \ref{subsection_discussion}.

\subsection{Knowledge Representation for Video Understanding} \label{Related_KR4VU}

Various approaches have been proposed to use knowledge representation for video understanding such as semantic-visual knowledge bases like FrameNet and Imagenet for modeling rich event-centric concepts and their relationships for video event detection \cite{IEEE_TMM2_KB_Video}, a knowledge and probabilistic driven framework for activity recognition \cite{IEEE_TMM4_KB_Video}, semantic representations for event detection \cite{IEEE_TMM5_KB_Video,IEEE_TMM6_KB_Video}.
Souza et al. deploy objects, actions and their bonds into graphs and use simulated annealing for event inference using temporal connections \cite{Souza1}, \cite{Souza3}. 
Ren et al. \cite{ren2013human} have previously proposed a Bayesian framework that utilizes object motions and their relations to improve object recognition reliability. 
This model enables robots to learn the interactive functionalities of objects from human demonstrations \cite{sun2014object} \cite{sun2015modeling}. 

Object information and analysis is an essential aspect for activity recognition.
The method in \cite{Reviewer3_3}, deploys spatial and functional constraints on the relationships between objects and motions to semantically interpret videos.
Modeling the mutual context of human pose and objects using a random field model \cite{Reviewer3_5}, modeling relationships between object parts and people in the scene using contextual scene descriptors and Bayesian learning \cite{Reviewer3_4}, and encoding objects for action classification and localization are examples of work on video understanding using object information.
These works all assume that a person is performing the act in the video therefore the human pose would be essential for their approaches.
We follow the path of incorporating objects and extend it to the goal of action recognition and activity inference by deploying our previously proposed knowledge representation network\cite{FOON}.
Our work is different from the mentioned object-based activity recognition methods in the manner that our videos do not contain a person and its pose. 
We only use the human hand and its location, if available in the scene, as a feature to  interpret video.

\section{Functional Object-Oriented Network} \label{section_overall_FOON}
The {\it functional object-oriented network}, FOON, is a knowledge representation for encoding knowledge about manipulation tasks and, in extension, object affordances.
A FOON can also be used by a robot for solving manipulation problems given a target goal. 
Currently, FOON focuses on learning activities in the cooking and kitchen domain, but it can also be extended to other domains and environments.

\subsection{FOON Basics} \label{section_FOON}
\begin{figure} [!h]
   \centering
   \includegraphics[width=8cm]{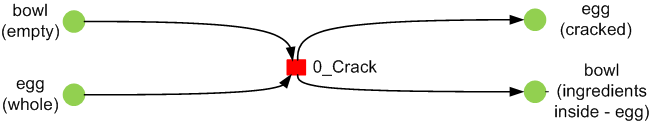}
   \caption{An example of a functional unit with 2 input object nodes (green), 2 output object nodes (green), and a motion node (red).}
\label{fig:example_fu}
\end{figure}

A FOON is a directed acyclic graph that contains two types of nodes (object and motion nodes), making it a bipartite network \cite{Networks_Int}. 
Figure \ref{fig:example_fu}
depicts a sample functional unit; the basic building block of a FOON. 

Object nodes are defined as items that are being manipulated or acted upon by a demonstrator, while motion nodes describe the action being applied on objects such as cutting, or mixing. 
An object node (${N_O}$) is identified by its {\it object type}, an {\it object state}, and a {\it motion identifier} which denotes whether the object is in motion during activity. 
Objects can also serve as {\it containers} of other objects, and each node can be described by a list of ingredients. 
Motion nodes are only identified by their {\it motion type}.
Within this graph, just like in regular bipartite networks, edges connect a pair of nodes; specifically, an edge in FOON connects an object-motion pair. 
The edge direction indicates the order in which an object may change in its state through a motion action similarly to Petri Nets \cite{Petri:2008} which require transitions to activate or fire place nodes. 

\subsection{Functional Unit}
A FOON consists of subcomponents or learning units called functional units. Each functional unit describes a single, atomic action as seen in an activity (an activity or subgraph can be considered as a series of actions). For instance, in the activity of cooking scrambled eggs, one functional unit may describe the action of cracking an egg, and another may describe the action of mixing the eggs in a bowl. A functional unit describes the transition of objects’ states before and after a manipulation motion occurs; this is described by input object nodes (objects before manipulation) and output object nodes (objects after manipulation). In this paper, our focus is generating these functional units directly from instructional videos for learning future instances of how tasks are executed.
A collection of subgraphs (or activities) that are merged together to combine knowledge and remove duplicate units is called a {\it universal} FOON.
Each functional unit has three components: input object nodes, output object nodes and a motion node that describes the action that possibly causes a change in the input objects' states. 
We say it possibly causes a state change because an action may not always incur a change of state. 
Each functional unit is also described by the time frames at which they are observed in an activity. 

\subsection{FOON Construction}
The graph shown in Figure 2 consists of nodes from 65 videos. These videos were annotated in the form of subgraphs, which consist of functional units that reflect each individual step in a cooking procedure. Edges would be drawn between an object node and motion node pair, where the object nodes are those seen in an action within the cooking activity and the motion node describes the action occurring. As we created these subgraphs and parsed them, we compiled a list of objects and motions to create labels for the different node instances seen and to enforce consistency in labels (as these subgraphs were created by multiple volunteers). When adding new information (subgraphs) from other datasets, we only need to annotate them to conform to the format of our graphs and parse them to get the labels correct. The merging procedure will add these newly parsed functional units to the network we have to ensure that there are no duplicates. This merging procedure is detailed more in our previous work in \cite{FOON}. This is where this proposed work fits in; the task of automatically generating subgraphs from videos (especially those from other datasets) is difficult to do and manual annotation can be time-intensive.

\subsection{FOON Sources and Statistics} 
A FOON ideally is learned directly from human demonstrations whether by video or from observation and it is automatically generated from such demonstrations. 
Although, in the earlier phases of constructing FOON, we opted to manually label YouTube videos as subgraphs. 
In the future, we will try to extend FOON using our method discussed in this paper.
After recording all functional units for a video, we parse the subgraph to ensure that all object and motion labels are consistent with all other subgraphs.

\begin{figure}[t]
  \centering
  \includegraphics[width=8cm]
{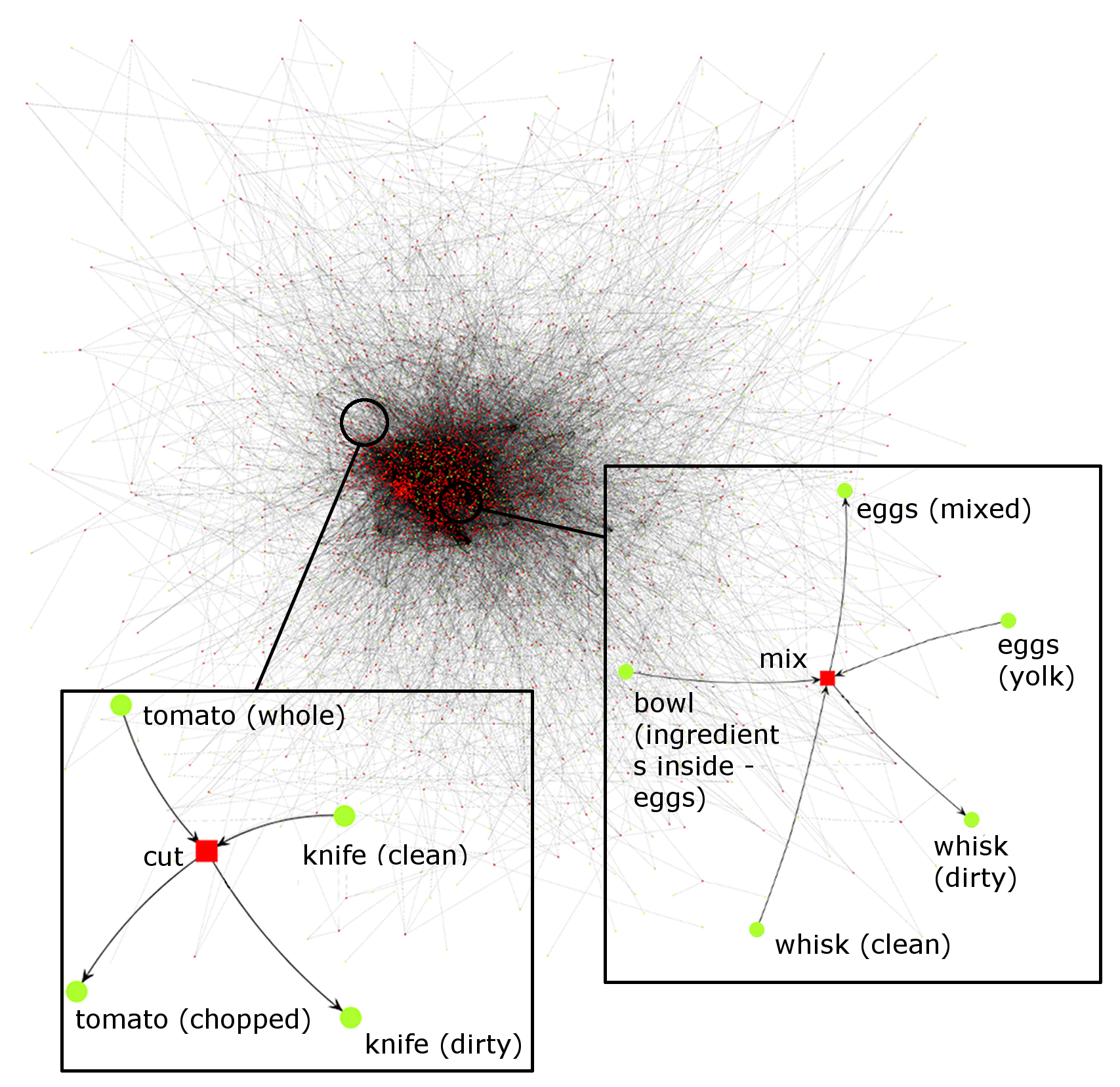}
  \caption{Illustration of an universal FOON with a total of 4955 nodes (both object and motion nodes). 
This FOON is comprised of many functional units such as those highlighted in the image.}
  \label{fig:uni_foon}
\end{figure} 

Each subgraph is then merged into a single network which we refer to as a {\it universal FOON}. 
The merging procedure is as follows: using a list of all functional units $G_{FOON}$, we compare each functional unit in all subgraphs to this list and append those units of a subgraph which are not present in $G_{FOON}$.

At the time of this paper, the universal FOON consists of data from 338 instructional videos, and a total of 3102 functional units.
This includes a subset of instructional videos from YouTube and videos from the MPII Cooking Activities Dataset \cite{MaxPlankIICooking}.
In total, the network contains 1853 object nodes, 3102 motion nodes, and 15656 edges. Figure \ref{fig:uni_foon} illustrates the network described by these statistics. 

\subsection{FOON vs Other Knowledge Representations}
Functional Object Oriented Network or FOON is not the first knowledge representation to address video understanding.
In this subsection we discuss the main differences between FOON and previous work.
Previous works in knowledge representation do not consider the joint representation of both objects and motions. Our work is inspired by the theory of affordance originally proposed in \cite{cite_affordances}. 
Many follow-up studies show that there is a link between manipulations and objects. Our objective is to create a graphical representation of manipulations where objects and motions describe affordance. 
In terms of graphical representations, previous works capture knowledge using probabilistic graphical methods or semantic graphs/trees. 
However, they do not create a knowledge base of activity from demonstrations which could then be used for performing (possible) new manipulations. 
In addition, for affordance studies, they would instead try to model the relationship between objects and simple actions to predict the effect or impact it has on them. A more general form of representation which is akin to FOON is Petri Nets, where place nodes are like object nodes and transition nodes are like motion nodes. Certain input places are needed to “fire” or execute a transition node much like input object nodes must be available to execute a given manipulation motion.

\section{The Video Understanding Pipeline} \label{section_pipeline} 
We propose a four-stage pipeline for video understanding. 
The pipeline identifies the objects and motions in a video sequence (associated with an action), and uses them together with the knowledge representation to assign a functional unit label to the event in action.
An {\it action} refers to a single, atomic event, and a {\it sequence} of actions represents an entire activity.
Consecutive identified actions will be analyzed as a whole to understand the activity (recipe) being executed in the video.
The steps to the pipeline are as follows 1) {\it functional object recognition}, 2) {\it functional motion recognition}, 3) {\it functional unit recognition}, and 4) {\it task graph inference}.

\begin{figure*} [!ht]
\centering
\includegraphics[width=0.75\textwidth]{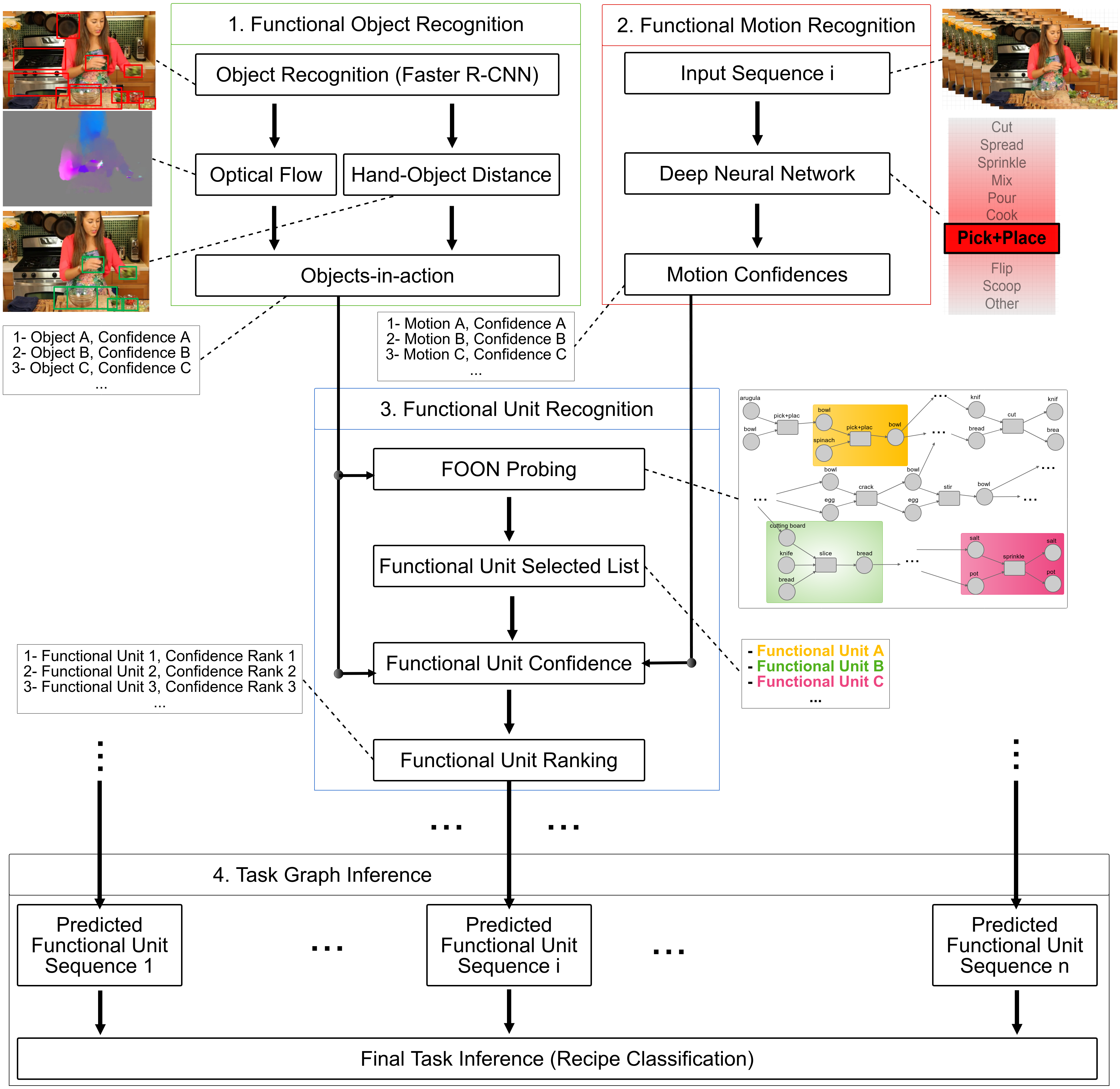}
\caption{The pipeline for automatic functional unit identification}
\label{fig:Pipeline}
\end{figure*}

In the first stage of the pipeline, the {\it functional object recognition} stage, all objects are identified and scores are assigned to objects based on their usefulness in the scene.
In the second stage, the {\it functional motion recognition}, each action (a split of the video) is classified into its corresponding motion class.
Using the results from these two stages and their FOON accordances, each action is analyzed and associated with a functional unit in the {\it functional unit recognition} stage.
The flow of recognized actions in video are analyzed and looked up in the FOON graph to classify them into an activity (recipe).
This last stage is referred to as {\it task graph inference}.
An illustration of the video understanding pipeline is depicted in Figure \ref{fig:Pipeline}.

\subsection{Functional Object Recognition} \label{section_local} 
We apply the well-known Faster R-CNN algorithm for localizing and labeling objects in the scene \cite{Faster_RCNN}. 
Faster R-CNN is a two-part convolutional network that detects object proposals and performs object classification simultaneously.
The output of the Faster R-CNN network is a set of bounding boxes and their corresponding object class labels.
We further identify the used objects in the video sequence, which we call objects-in-action, using three metrics namely the closeness of the human hand to the object, the magnitude of flow and the frequency in which the objects have been observed in the video.
We explain the functional object recognition stage more thoroughly in Section \ref{section_inAction}.

\subsection{Functional Motion Recognition} \label{subsection_motion}
In some cases FOON is not able to correctly identify the action in video using only object features. 
For example, knowing that the objects "bowl" and "egg" are objects-in-action could lead to multiple FOON inferences. 
Because various functional units contain the object nodes "bowl" and "egg" but have different motion nodes (e.g. pouring or cracking).
In another example, when sprinkling salt with the hand, it is difficult to visually discern that the object "salt" is being used, but the hand motion will suggest the action of sprinkling.

To address these issues, we fine-tune the deep (CNN+LSTM) network by Donahue et al. \cite{CNN+LSTM} with 10 classes in the last layer.
This network comprises a CNN portion and an LSTM portion.
The frames of a sequence are one by one given as input to the CNN and the output of the CNN is given as input to the LSTM layer. 
The outputs of the LSTM layer is averaged upon to provide a final prediction for the class of the motion in action.
The architecture of the CNN network comprises of five convolutional layers and two fully connected layers. 
The initial five convolutional layers and a single fully connected layer on top is fed to one layer of a recurrent LSTM layer. 
The output of the LSTM layer is followed by the classification layer. 
We modified the last layer so that the number of neurons in the last layer of the network contains ten neurons to reflect the ten motion types we have picked for training. 
We train the CNN architecture and the CNN + LSTM architectures separately. 
We use the trained weights from \cite{CNN+LSTM} and only perform training for the last layer of classification.
We only report the better results from the CNN+LSTM architecture.

Each motion class in the deep model is associated with a set of motion nodes in FOON.
The network assigns confidence scores to each of the motion classes. A confidence score reflects the probability of a class being assigned as the label for the action happening in video.
For more details on the approach, we refer readers to the algorithm described in \cite{CNN+LSTM}. 
The output from this deep network is used to calculate confidences for each candidate functional unit in the functional unit recognition stage.

\subsection{Functional Unit Recognition} 
Objects-in-action are looked up in the universal FOON to identify candidate functional units.
Candidate functional units are evaluated based on a confidence score which is calculated in this stage an is thoroughly discussed later in the paper. 
This consolidated confidence score incorporates both object confidences produced from the functional object recognition stage and motion confidences resulted from the functional motion recognition stage.
The confidence score estimates how related each candidate functional unit is to the on-going action in the present sequence.
The list of candidate functional units is further sorted based on their confidences. Functional units with the highest confidences are associated with the current action.

\subsection{Task Graph Inference} 
To identify the activity (sequence of actions) in a video, the identified actions throughout the video are used together with FOON look-up to predict the most likely activity label for that video.

\section{Functional Object Recognition} \label{section_FUA} 
We recognize and localize all objects in a video sequence (associated with an action) using the well-known Faster R-CNN algorithm \cite{Faster_RCNN}. 
We then quantify the involvement of each object in the current action, by extracting optical flow features and calculating hand-object distances in each frame of the video sequence.
A list of the most used objects is created based on the extracted features. 
We name the list as objects-in-action.

\subsection{Recognizing Objects-in-action} \label{section_inAction} 
In this stage of the pipeline we use the bounding box associated with each object for our computations.
After localizing objects, the less frequent objects in the video are excluded.
The center point of the bounding boxes resulted from the Faster R-CNN algorithm are used to calculate the {\it object's average distance from the hand}.
The distances are further normalized using a Gaussian distribution. 
The {\it optical flow of objects} within the video sequence are exploited.
The proposed method in paper \cite{OpticalFlow} is used to estimate the optical flow between two frames. 
The estimated optical flow and the objects' positions are incorporated to estimate the flow of each object. 
Objects with higher magnitude of flow are assigned a higher confidence value. 
A higher value conveys a higher chance that the object is moving and hence a higher probability that the object is being used in the video sequence.
Equation \ref{equation_2} shows how these metrics are integrated to estimate a confidence for each object.

\begin{equation}
\label{equation_2}
conf_{object} = {\alpha}.c_{flow}+{\beta}.c_{dist}+{\gamma}.c_{freq}`'
\end{equation}

\begin{figure} [!ht]
\centering
\includegraphics[width=0.35\textwidth]{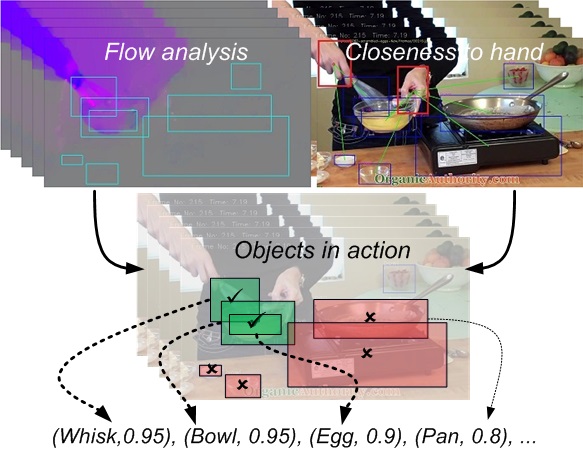}
\caption{An example showing the procedure of identifying objects-in-action. 
Items such as "egg" and "whisk" would be possible candidates for participating in an egg whisking motion.}
\label{fig:objects_in_action}
\end{figure}

In Equation \ref{equation_2}, \(c_{flow}\), \(c_{dist}\) and \(c_{freq}\) are the optical flow confidence, distance to hand confidence, and frequency confidence of each object respectively; $conf_{Object}$ is the final calculated confidence of the object. 
Coefficients \({\alpha}\), \({\beta}\) and \({\gamma}\) are tuned manually and represent how much each factor contributes to the final confidence of each object.
Figure \ref{fig:objects_in_action} depicts the procedure of identifying objects-in-action for a simple action of whisking eggs, using the three metrics mentioned in equation \ref{equation_2}.

In this example, we observe an egg whisking motion happening in which objects "egg", "whisk" and "bowl" are in the top of the list of objects-in-action and objects "pan" and "stove" have lower confidences.
 
\section{Functional Unit Recognition}  \label{section_FUR} 
Each action in video is associated with the closest functional unit from FOON. 
To associate the correct functional unit with an action, unrelated functional units are filtered out.
Filtering is performed using {\it functional unit confidence} estimation, and {\it probing} which we will discuss in this section.

\subsection{Functional Unit Confidence} \label{section_confidence}
The pipeline recommends a list of in-use objects from the current action named objects-in-action (Section \ref{section_inAction}).
Objects from the list are looked up in FOON, and functional units containing them are identified.
The identified functional units are suggested as candidate functional units that can be associated with the current action in the video.
Every functional unit contain several object nodes which may or may not be included in the list of objects-in-action.
The overlap between the object nodes (of a functional unit) and the objects-in-action is called the used set and the remainder of the object nodes is called the unused set.
These two sets of objects are used to determine whether we should support or penalize a candidate functional unit.
Equation \ref{eq:foon_conf} shows how the confidence of a candidate functional unit is estimated.

\begin{equation}
\label{eq:foon_conf}
conf_{FOON} = \frac{\sum_{n=1}^{N_{used}} conf_{n}}{N_{used}} - penalty + {\kappa}.bonus
\end{equation}

In this equation, \(conf_{FOON}\) is the estimated confidence, \({N_{used}}\), is the number of object nodes in the used set of a candidate functional unit, and \(conf_{n}\) is the confidence of each of those objects (subsection \ref{section_inAction}). 
The \(bonus\) term is estimated based on the pixel-wise overlap of all objects used in a candidate functional unit.
This term represents the extent of interaction between  the objects.
The \(penalty\) term calculated by Equation \ref{equation_4}, represents the penalty applied to the estimated confidence.

\begin{equation}
\label{equation_4}
penalty = \sum_{m=1}^{N_{notused}} {\lambda}.conf_{m} + \sum_{k=1}^{N_{extra}} {\eta}.conf_{k}
\end{equation}

\begin{figure} [!ht]
\centering
\includegraphics[width=0.4\textwidth]{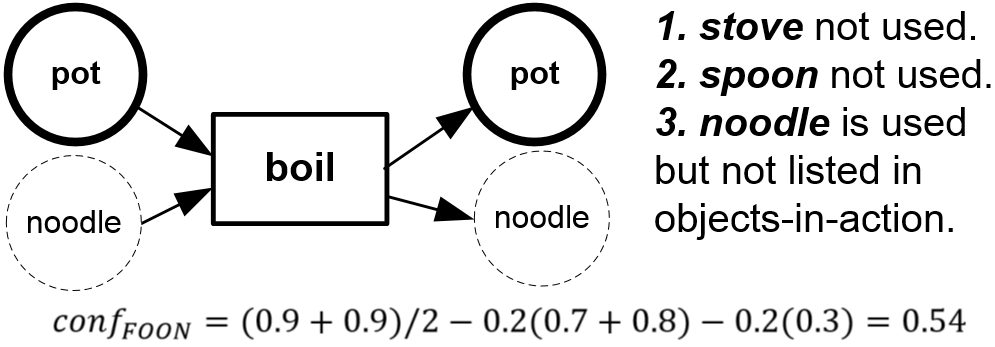}
\caption{Illustration of functional unit confidence estimation. In this example, identified objects-in-action are Pot, Spoon and Stove with confidences 0.9, 0.8, and 0.7 respectively ($\lambda$=$\eta$=0.2).}
\label{fig:confidence}
\end{figure}

The confidence of the objects listed in the list of objects-in-action but not used in the candidate functional unit, \(conf_{m}\), together with the confidence of the objects not listed as objects-in-action but used in the candidate functional unit \(conf_{k}\), contribute to the penalty. 
In this equation, \({N_{notused}}\) is the number of unused objects, and \({N_{extra}}\) is the number of objects not listed but used in the candidate functional unit. 
In Equation \ref{eq:foon_conf}, the constant \({\kappa}\) tunes the effect of bonus and penalty. 
The constant \({\lambda}\) in Equation \ref{equation_4} tunes the effect of unused objects on the penalty term and the constant \({\eta}\) tunes the effect of objects used but not listed.
Figure \ref{fig:confidence} illustrates the procedure of confidence estimation for a candidate functional unit.

The confidence calculated in Equation \ref{eq:foon_conf} focuses solely on object interaction and their functional accordances.
We believe motion can introduce additional information for confidence calculation.
To include functional motion for estimating the confidence, we fuse the output of the CNN+LSTM network with the confidence estimated solely based on object interaction, \(conf_{FOON}\).
The output of the CNN+LSTM network for motion recognition has 10 confidence scores representing the probability of each of the motion classes happening. 
We rank the motion classes based on their resulted confidence scores. 
Finally, the confidences of functional units in Equation \ref{eq:foon_conf} are fused with the results from the CNN+LSTM network to extract a final confidence for the functional units as shown in Equation \ref{eq:motion_conf}.

\begin{equation}
\label{eq:motion_conf}
conf_{motion} = conf_{FOON} + {\alpha}.conf_{LSTM}
\end{equation}

In Equation \ref{eq:motion_conf}, \(conf_{FOON}\) is the confidence calculated in equation \ref{eq:foon_conf}, and \(conf_{LSTM}\) is the confidence calculated based on results from the CNN+LSTM network. Coefficient \(\alpha\) balances the effect of each of those parameters.

\subsection{Probing}\label{ind_probing}
Each object is individually looked up in FOON and all functional units containing that object are identified.
A list of candidate functional units containing the object is acquired.
The list contains candidate functional units that may associate with the current action.
We exclude the objects with lower confidences, \(conf_{object}\), from the list, to reduce the number of potential objects-in-action and in consequence, the number of probed objects and candidate functional units.
To illustrate an example, let us assume that the filtered list of objects seen in the sequence or in other words probed objects are (egg, bowl, and fork) and the ground truth functional unit associated with the sequence has the motion node "mix" with the objects "bowl", "egg", and "fork" as input nodes and "egg", and "fork" as output nodes.
Individually probing functional units in FOON using the list of probed objects produces a list of candidate functional units that contain those objects. 
Table \ref{table_1} shows some of the 674 candidate functional units that contain the probed objects for this specific example. 
Any other functional unit that is not identified does not contain the probed objects. 

\begin{table} [h]
\centering
\caption{Results of probing objects in FOON based on the example in Section \ref{ind_probing}. The objects-in-action are shown in bold.}
\label{table_1}
 \begin{tabular}{|@{\hskip4pt}c@{\hskip4pt}|@{\hskip4pt}c@{\hskip4pt}|@{\hskip4pt}c@{\hskip4pt}|@{\hskip4pt}c@{\hskip4pt}|@{\hskip4pt}c@{\hskip4pt}|}
 \hline
 & Input Nodes & Motion & Output Nodes & Overlap \\ [0pt]
 \hline
 1 & mixer, \textbf{bowl} & mix & mixer, \textbf{bowl} & 0.5 \\ [0pt]
 \hline
 \rowcolor{gray}
 2 & \textbf{fork}, \textbf{egg}, cup & stir & \textbf{fork}, \textbf{egg}, cup & 0.67 \\ [0pt]
 \hline
 : & : & : & : & : \\ [0pt]
 \hline
 674 & \textbf{bowl}, pan, pasta & pour & pan & 0.25 \\ [0pt]
 \hline
\end{tabular}
\end{table}

Each probed functional unit from FOON contains object nodes that may or may not have been seen in the current video sequence (associated with an action).
The last column of Table \ref{table_1} depicts the overlap between the objects included in a probed functional unit with the identified objects in the video sequence.
The probed functional units with an overlap value less than a specific threshold are excluded.
Confidence values for the remaining functional units are calculated and the ones with the highest confidence values are selected.
The selected functional units are the most likely to be associated with the on-going action.

\section{Experiments and Results} \label{section_Exps} 
In our experiments, we use the annotated videos that were used for the creation of the universal FOON in \cite{FOON} and the videos from the MPII Cooking Activities Dataset summing up into a total of 338 videos \cite{MaxPlankIICooking} \footnote{The videos and graphs of FOON are available at: \href{http://www.foonets.com/}{http://www.foonets.com}}.
For our current experiments, we also manually label some of the video sequences in FOON with object bounding boxes and their categories.
We use 11 of the 338 cooking videos as our test dataset, which includes an overall amount of 55 functional units.
We perform our tests in 11 iterations in a leave-one-out manner: in each iteration, one video is entirely left out, while the rest of the videos are used to create a FOON. 
The lack of training data for object recognition, lack of labeled ground truth data for the videos, and a low number of instances of every kind of functional unit in the dataset forced us to only use 11 videos for our experiments. 

For testing the pipeline, we conducted three different experiments based on both manually and automatically labeled objects. These experiments are as follows: 1) Comparing functional unit recognition using only FOON look-up with functional unit recognition using the fusion of FOON and motion recognition. 2) Video understanding for functional unit recognition with and without FOON. 3) Task inference and recipe classification. 

\subsection{Object Overlap Metric} \label{subsection_overlap}
\begin{figure} [!htb]
\centering
\includegraphics[width=0.35\textwidth]{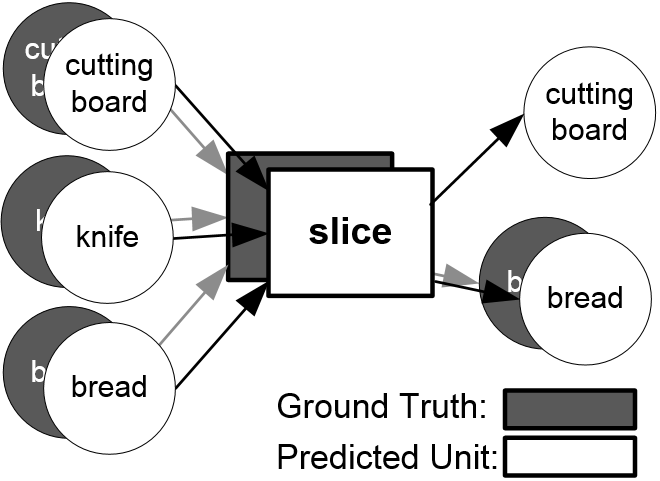}
\caption{
An illustration of how overlap is measured. 
In this example, overlap is equal to 4 nodes, precision is 100\% and recall is 80\%. 
If the ground truth motion node was anything but "slice", precision and recall would be 0\%.}
\label{fig:overlap}
\end{figure}

The overlap between a candidate functional unit and its corresponding ground truth functional unit is used to evaluate the results.
This overlap metric is calculated for each action in the video separately.
The metric we use is fairly simple: if the motion node of the candidate is equivalent to the motion node of the ground truth, the overlap between their object nodes is counted.
Consequently, precision and recall are computed using the object overlap.
{\it Precision} is measured as overlap divided by the number of object nodes in the candidate, while {\it recall} is measured as overlap divided by the number of object nodes in the ground truth. 
If the motion nodes are different, precision and recall are assumed to be 0. Figure \ref{fig:overlap} illustrates how precision, and recall are calculated. 
 
\subsection{Functional Unit Recognition}
We use the time stamp labels in the universal FOON, to split the videos in the dataset into its comprising actions.
For example, a video demonstrating a cook making scrambled eggs would be split into several atomic actions such as cracking the eggs, pouring the eggs into a bowl, and mixing the eggs with a whisk. 

\subsubsection{Functional Unit Recognition using FOON}
\begin{figure} [!h]
\centering
\includegraphics[width=0.5\textwidth]{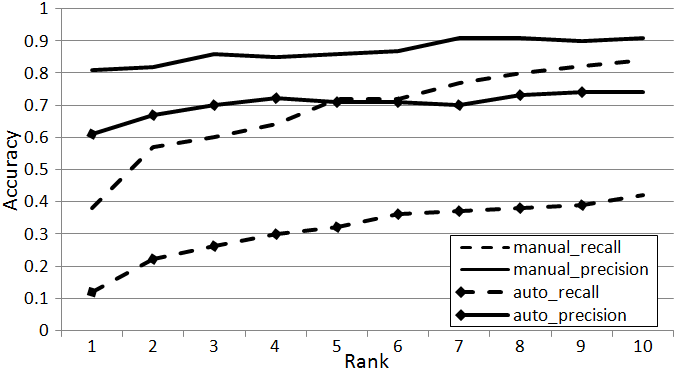}
\caption{Precision and recall as observed in manual and automatic object recognition for top 1 to 10.}
\label{fig:accuracy_rank}
\end{figure}

Each action sequence in a video is fed to the algorithm that identifies the best functional unit fitting the action based on the metrics discussed in Section \ref{section_FUA}.
In each iteration, we use a single video for evaluation, and the other 337 videos, to create an iteration-specific FOON. 
To identify the functional unit corresponding to an action sequence in a video, the sequence is processed based on the iteration-specific FOON.
After identifying functional units, precision and recall are computed as defined in Section \ref{subsection_overlap} for all candidate functional units for top 1 to 10 results as shown in Figure \ref{fig:accuracy_rank}.

\begin{figure*} [!htb]
\centering
\includegraphics[width=0.9\textwidth]{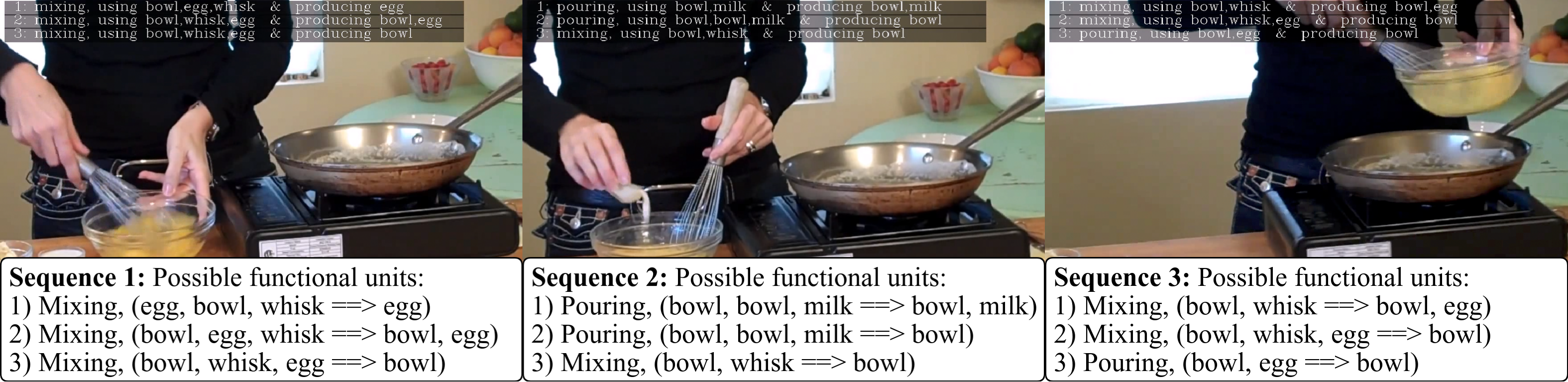}
\caption{An example of functional unit recognition using labeled and manually split sequences for the scrambled egg recipe.}
\label{fig:FU_Recognition_MM}
\end{figure*}

In Figure \ref{fig:accuracy_rank}, the horizontal axis represents the number of best functional units analyzed for precision and recall calculation.
The solid curves show precision and the dashed curves show recall calculated on 55 functional units for both manually  and automatically labeled objects.
Figure \ref{fig:accuracy_rank} shows that the algorithm can potentially improve with additional procedures.
We can also see that precision in Figure \ref{fig:accuracy_rank} is always higher than 80 percent, showing that our algorithm sometimes misses the objects in the video; however, when it assumes an object is being used in a functional unit, it usually identifies the functional unit correctly.
In Figure \ref{fig:FU_Recognition_MM} snapshots of three sequences of a cooking video is depicted with their predicted functional units.
In this example, the correct functional unit is always included in the top three identified functional units.

\subsubsection{Functional Unit Recognition with Motion Recognition and FOON}
We fuse motion recognition with FOON look-up to improve the recognition procedure. 
We create motion classes by selecting the 9 most frequent motion types from the FOON motion nodes (e.g. "pour", "pick+place", and "cook") \cite{FOON}. 
To accommodate the other types of motions, we design a class labeled as the "other" class. 
We extract flow features from each sequence in the video.
We apply the CNN+LSTM network on RGB and flow sequences of each event and perform an averaging of the outputs from the two networks.
The architecture returns 10 values representing confidences for the 10 classes. The motion confidence values are used in the calculation of candidate functional unit confidences.
Table \ref{table_6} shows the top 1, top 3, top 5, and top 10 accuracy of prediction for functional unit recognition using both FOON and motion recognition.
  
\begin{table} [!ht]
\centering
\caption{Top 1 to 10 accuracy of prediction for functional unit recognition using FOON and Motion Recognition.}
\label{table_6}
 \begin{tabular}{|c |c |c |}
 \hline
\textbf{} & \textbf{Using FOON} & \textbf{Using FOON + Motion Recognition}\\ [0pt]
 \hline
 Top 1  & 56\% & 64\% \\ [0pt]
 \hline
 Top 3  & 75\% & 84\% \\ [0pt]
  \hline
 Top 5  & 80\% & 89\% \\ [0pt]
  \hline
 Top 10  & 89\% & 98\% \\ [0pt]
 \hline
\end{tabular}
\end{table}

The accuracy of prediction for an action is computed by comparing the identified functional units with the ground truth functional units. 
If the motion node of the identified functional unit is equivalent to the motion node of the ground truth and the overlap of object nodes is higher than 80\% we conclude the prediction as correct.
We count the number of correct predictions over all functional units in the test set and calculate the accuracy.
In some cases the motion node of the ground truth may vary in text with the motion node of the identified functional unit, while having the same interpretation (e.g. "whip" vs "stir", or "slice" vs "cut"). 
These cases of motion nodes are considered equivalent.

As shown in Table \ref{table_6}, the accuracy of functional unit recognition when motion recognition is fused with FOON look-up is higher than functional unit recognition without motion recognition.
This shows adding automatic motion recognition to the pipeline improves the motion node recognition and leads to better identification of functional units.
The deep network guesses the motion node in only 47 percent of the cases. 
The complexities of the videos such as background variations, different camera views, and moving cameras prevent it from producing the desired accuracy.
In experiments, we set \(\alpha\) in Equation \ref{eq:motion_conf} to less than 0.2, so the results from the neural network would not adversely influence the final results.

\subsubsection{Analysis}
To see the effect of each part of the pipeline on the results, we look deeper into each part.
The automatic motion recognition by itself achieves 67 percent accuracy, while the functional unit recognition without motion recognition achieves 61 percent accuracy (top 2).
There are two differences in these two evaluations that make them incomparable.
First, for automatic motion recognition, the number of classes of motion is generalized and reduced to 10 classes, while for functional unit recognition, there are over 50 types of motion nodes. 
Second, functional unit recognition identifies the action with focus on both the objects and the motion occurring, while the aim of motion recognition is to recognize the motion class in an action. 
Although they are not comparable, motion recognition is a good feature to fuse with FOON for optimal functional unit recognition.

We calculate the overlap between objects-in-action and the identified functional units as 84 percent.
This shows that although the majority of objects have been identified correctly, the accuracy of functional unit recognition is lower than expected due to mistakes in identifying the motion nodes. 

\begin{figure*} [!htb]
\centering
\includegraphics[width=0.8\textwidth]{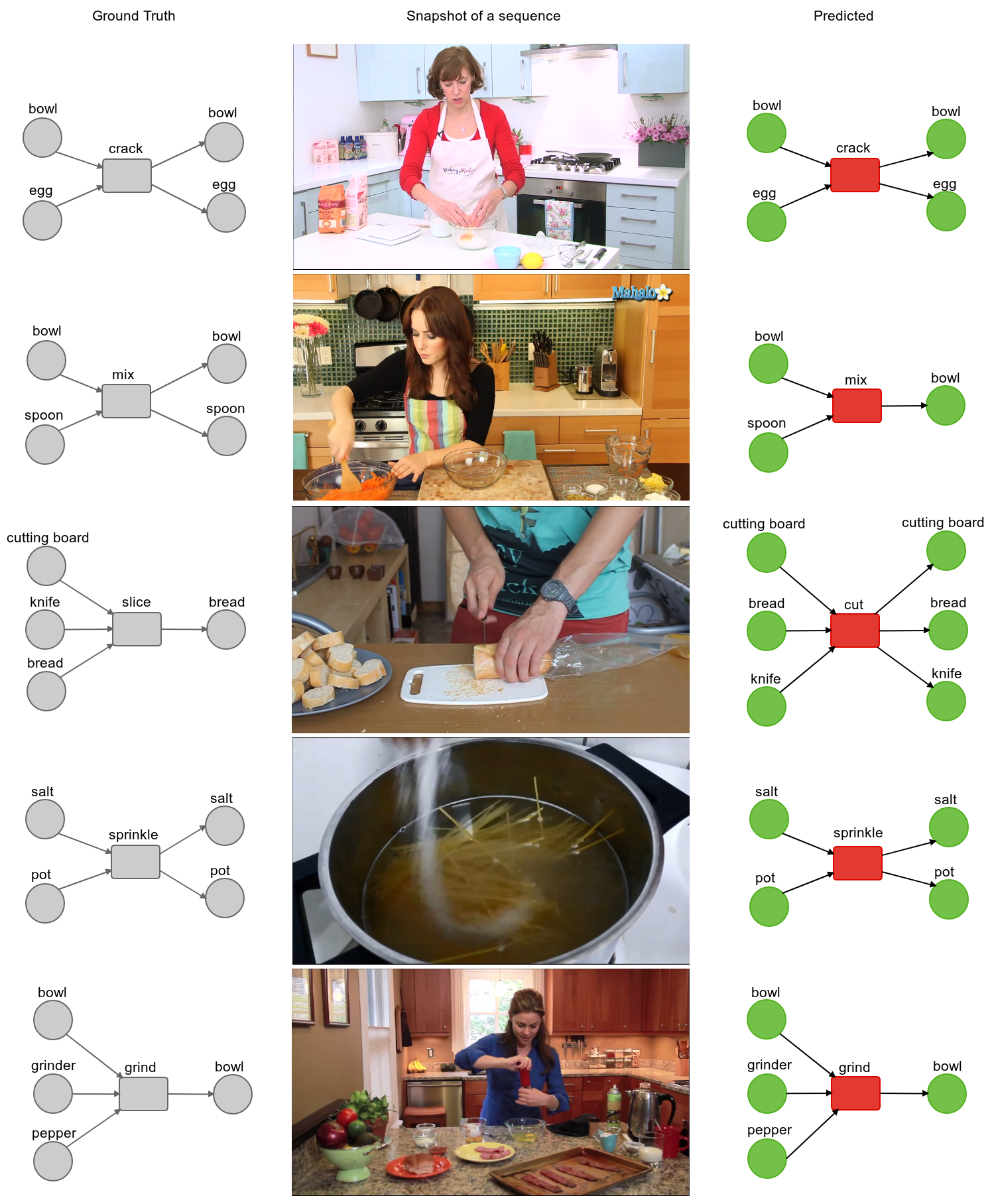}
\caption{Snapshots of events, their ground truth functional unit representation and the predicted functional unit for them.}
\label{fig:qualitative}
\end{figure*}

In another experiment, we applied the pipeline fused with motion recognition for automatically recognized objects and report its top 1 to 10 results in Figure \ref{fig:accuracy_rank}.
Although, object recognition is an important stage of the pipeline that can be improved, we do not address it further, as it is not our specific goal in this paper.
Snapshots of various sequences with their ground truth representation and identified functional units is depicted in Figure \ref{fig:qualitative}.

\subsection{Video Understanding}
\begin{figure} [!htb]
\centering
\includegraphics[width=0.4\textwidth]{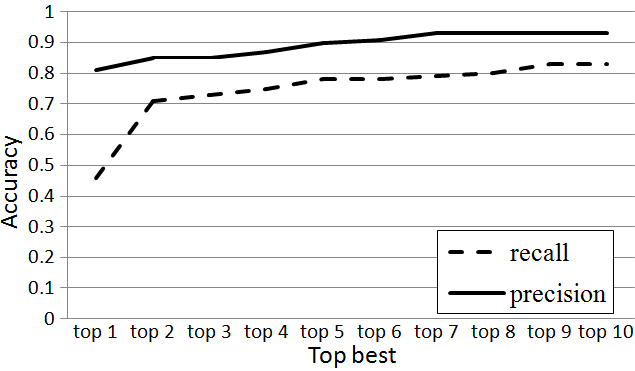}
\caption{Graph showing the results of the overlap metrics for video understanding for top 1 to 10.}
\label{fig:accuracy_video}
\end{figure}

The pipeline is evaluated based on the extent it understands a video using the overlap metric.
Precision and recall is calculated for both object and motion nodes for all actions of each video individually, and the average precision and recall is calculated for all videos over the top 10 results. 
Figure \ref{fig:accuracy_video} shows the calculated results.

The results show that the pipeline is capable of perceiving an understanding of the video especially when top 5 results are used.
The lower values for recall may be due to the errors made in identifying objects-in-action.
We calculate the F-Score metric using recall and precision as discussed in \cite{F_Score}.

\begin{figure} [!htb]
\centering
\includegraphics[width=0.4\textwidth]{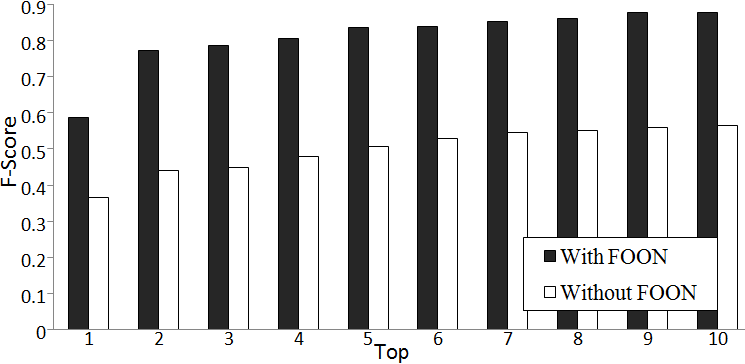}
\caption{Graph showing the calculated F-Scores for Video Understanding with and without FOON.}
\label{fig:accuracy_Foon_NoFoon}
\end{figure}

The video understanding F-Score is calculated for our pipeline in two instances: 1) when FOON is used, and 2) when FOON is not used and the results are depicted in Figure \ref{fig:accuracy_Foon_NoFoon}.
When using FOON, we calculate the F-Score by using the overlap metric for ground truth and identified functional units. 
When not using FOON, we calculate the overlap metric between the highest ranked objects and the objects in the ground truth functional unit. 
We calculate the overlap between the highest ranked motion classes with the motion nodes in the ground truth. 
The sum of these two overlaps is used to calculate the precision and recall and F-Score.
Using FOON achieves higher F-Scores than not using FOON since object and motion nodes in a video are perceived much better when using FOON as reference.

\subsection{Task Inference and Recipe Classification}
We deploy our algorithm for recipe classification of unseen cooking videos. 
We use 8 videos including 1 salad recipe, 2 omelette recipes, 2 bread recipes, 1 cake recipe, 1 noodle recipe, and 1 sandwich recipe for the test.
We classified all the recipes in FOON into 13 classes of recipes namely: cake, pizza, bread, omelette, soup, barbecue, sandwich, smoothies, pasta, coffee and tea, salad, mashed potato, and others. 

\begin{figure} [!ht]
\centering
\includegraphics[width=0.45\textwidth]{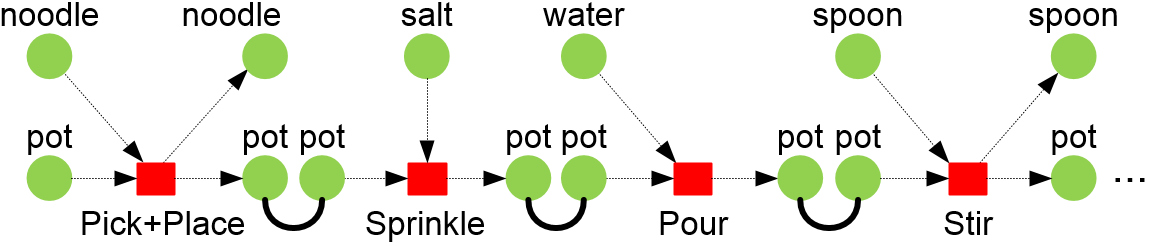}
\caption{An illustration of identified functional units for the noodle cooking video (identified as noodle by the proposed pipeline)}
\label{fig:noodle_recipe}
\end{figure}

Task inference is performed after all functional units in a video are identified. 
All identified objects-in-action that are used in the video and identified functional units equally contribute to the task inference stage.
To classify a video to a recipe, clusters of recipes are created using all videos in the train set.
The similarity distance between the current video and each cluster is calculated and the closest cluster is selected as the recipe associated with the video. 
To calculate the similarity distance between the current video and a cluster, the similarity with each of the videos in the cluster is calculated and is averaged.
Similarity distance between a video and a recipe is calculated as the similarity of functional units in the video with the similarity of functional units in the recipe aggregated with the similarity of used objects in the video with the similarity of object nodes in the recipe.
In our similarity comparison we do not check the order of functional units.
The recipe class with the highest similarity is assigned to the video.
Figure \ref{fig:noodle_recipe} shows the identified functional units of a video demonstrating a cook making noodles. 

We report top 1 and 2 results of recipe classification in Table \ref{table_7}.
The recipe classification algorithm returns the predicted class names based on their confidence scores. 
If the class name with the highest confidence is the same as the ground truth class name, the classification is correct. 

\begin{table} [h]
\centering
\caption{Recipe classification results.}
\label{table_7}
 \begin{tabular}{|c |c |c |} 
 \hline
Used Procedure & Top 1 & Top 2 \\ [0pt]
 \hline
 Manually labeled Objects  & 37.5\% & 100\% \\ [0pt]
 \hline
 Automatically labeled Objects  & 25\% & 75\% \\ [0pt]
 \hline
\end{tabular}
\end{table}

As shown in Table \ref{table_7}, the algorithm using FOON can approximately guess what recipe is being cooked in the video granted that all objects in the video sequence are identified correctly.
The motion of the objects can also insinuate the type of recipe activity that is happening.

\subsection{Discussion} \label{subsection_discussion}
There are different literature that work on activity recognition using either knowledge bases or other methods, but they represent a video with a sentence or a label for the activity. Our work outputs sub-graphs representing short activities for each part of the video. This makes our work incomparable to other work. Therefore we analyze our work through the overlap metric and compare two approaches for video understanding; the pipeline using FOON and the pipeline not using FOON. 
It is clear that some methods in the literature can be substituted with the method we use to integrate with FOON, but our current focus is to prove that FOON is a powerful knowledge representation that can understand video and would be able to semi-automatically build itself in the future.

The proposed framework focuses on understanding events and tasks in single camera videos.
However, due to the importance of multi-view applications,
we discuss a few ways that the framework can be integrated into a multi-view system.
The proposed framework can individually be applied to multiple videos in a multi-view system. 
Individual predictions can be gathered from multiple deployments of the framework. 
The predictions will further be combined to reach to a final prediction of the actions and activity in the video.
We can also fuse multiple views at the confidence level.
Confidences of objects can be extracted at each view, and fused to reach a final confidence for the objects.
The framework can further run as proposed.

The goal of the proposed framework is to identify the events and task in a video.
The framework can be used as the vision system of a robot chef, or in any robotic system that deploys and manipulates utensils, such as a robot carpenter, robot waiter, etc.

\section{Conclusion and Future Work} \label{section_Conclusion} 
The main objective of the paper is video understanding with the help of the FOON knowledge representation.
We proposed a pipeline for video understanding using the functional object-oriented network (FOON) \cite{FOON} and deep neural networks.
We make use of low-level image features together with deep networks to identify objects of interest. Using objects of interest (which we call objects-in-action) and deep motion understanding, we associate the actions in a video with the correct functional units in the knowledge representation (FOON). 
We demonstrated that using FOON significantly improves the performance of video understanding in comparison to not using FOON.

Our current pipeline is a big step towards automatically extending the knowledge representation graph. 
Automatically extending the graph would present a massive improvement to the network's applications such as robots solving manipulation problems given a target goal. 
In future work, we would like to explore other methods of identifying objects-in-action, incorporate object recognition confidences to handle mis-identified objects and incorporate history of events using FOON for inference. 
We are also working on generalizing the knowledge contained within a FOON to achieve more generic inferences from FOON.

\ifCLASSOPTIONcaptionsoff
  \newpage
\fi

\section{Acknowledgments} \label{section_ack}

This material is based upon work supported by the National Science Foundation under Grants No. 1421418 and No. 1560761.

{\small
\bibliographystyle{unsrt}
\bibliography{references}
}

\end{document}